\documentclass[lettersize,journal]{IEEEtran}
\usepackage{amsmath,amsfonts}
\usepackage{algorithmic}
\usepackage{array}
\usepackage[caption=false,font=footnotesize,labelfont=rm,textfont=rm, subrefformat=parens]{subfig}
\usepackage{textcomp}
\usepackage{stfloats}
\renewcommand\ref[1]{\subref*{#1}}
\usepackage{url}
\usepackage{verbatim}
\usepackage{graphicx}
\usepackage{booktabs}
\usepackage{amssymb}
\usepackage{mathtools} 
\usepackage{amsmath}
\usepackage{multirow}
\usepackage{graphicx}
\usepackage{booktabs}
\usepackage{color}
\usepackage{hyperref}
\graphicspath{{images/}}
\graphicspath{{line_graph/}}
\graphicspath{{curve/}}
\graphicspath{{heatmap/}}
\graphicspath{{authors/}}
\usepackage[ruled,linesnumbered]{algorithm2e}
\def\BibTeX{{\rm B\kern-.05em{\sc i\kern-.025em b}\kern-.08em
    T\kern-.1667em\lower.7ex\hbox{E}\kern-.125emX}}

\usepackage{atbegshi}

\begin{document}
\AtBeginShipoutNext{\AtBeginShipoutDiscard}

\title{Enhancing Road Safety through Accurate Detection of Hazardous Driving Behaviors with Graph Convolutional Recurrent Networks}

\author{
  \href{https://orcid.org/0000-0002-4968-5800}{\includegraphics[scale=0.06]{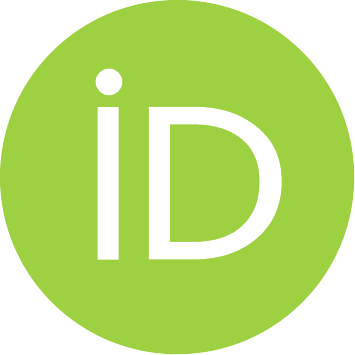}\hspace{1mm}Pooyan Khosravinia},
  \href{https://orcid.org/0000-0003-0698-5413}{\includegraphics[scale=0.06]{orcid.pdf}\hspace{1mm}Thinagaran Perumal},
  \href{https://orcid.org/0000-0002-1558-1907}{\includegraphics[scale=0.06]{orcid.pdf}\hspace{1mm}Javad Zarrin}}
  
 \thanks{
 P. Khosravinia, and T. Perumal are with the Faculty of Computer Science and Information Technology, Universiti Putra Malaysia (UPM), 43400 Serdang, Malaysia).}
 \thanks{
 J. Zarrin is with the School of Design and Informatics (SDI), Abertay University, Bell Street, Dundee DD1 1HG, UK (e-mail: j.zarrin@ abertay.ac.uk).}

\maketitle
\thispagestyle{empty}
\newpage
\setcounter{page}{1}

\begin{abstract}
Car accidents remain a significant public safety issue worldwide, with the majority of them attributed to driver errors stemming from inadequate driving knowledge, non-compliance with regulations, and poor driving habits. To improve road safety, Driving Behavior Detection (DBD) systems have been proposed in several studies to identify safe and unsafe driving behavior. Many of these studies have utilized sensor data obtained from the Controller Area Network (CAN) bus to construct their models. However, the use of publicly available sensors is known to reduce the accuracy of detection models, while incorporating vendor-specific sensors into the dataset increases accuracy. To address the limitations of existing approaches, we present a reliable DBD system based on Graph Convolutional Long Short-Term Memory Networks (GConvLSTM) that enhances the precision and practicality of DBD models using public sensors. Additionally, we incorporate non-public sensors to evaluate the model's effectiveness. Our proposed model achieved a high accuracy of 97.5\% for public sensors and an average accuracy of 98.1\% for non-public sensors, indicating its consistency and accuracy in both settings. To enable local driver behavior analysis, we deployed our DBD system on a Raspberry Pi at the network edge, with drivers able to access daily driving condition reports, sensor data, and prediction results through a monitoring dashboard. Furthermore, the dashboard issues voice warnings to alert drivers of hazardous driving conditions. Our findings demonstrate that the proposed system can effectively detect hazardous and unsafe driving behavior, with potential applications in improving road safety and reducing the number of accidents caused by driver errors.
\end{abstract}


\section{Introduction}
\label{sec:introduction}
Unsafe driving is an abnormal pattern of social behaviors such as speeding, tailgating, rude gesturing, honking, improper lane changing, and distracted driving, which constitute a serious threat to public safety \cite{article033}. According to data collected in the year 2019 from the AAA foundation, approximately 80 percent of drivers in the United States of America engaged in unsafe driving behaviors at least once in the 30 days before the survey \cite{TheAAAFoundationforTrafficSafety2019}. Evidently, Driving behavior is one of the important factors involved in driving safety \cite{Chen2019}. Many studies have implemented to develop a driving behavior detection (DBD) system which can detect the drivers' actions and help them to drive safely. Most studies used machine learning or deep learning algorithms to solve this problem. The support vector machines (SVM), hidden Markov models (HMM), k-nearest neighbors (KNN), artificial neural networks (ANN), recurrent neural networks (RNN), and convolutional neural networks (CNN) are the common algorithms used in previous papers for building a model that can detect the driving behavior accurately and efficiently.

The collection of driving behavior data is an important aspect of a DBD system. Currently available techniques for monitoring and collecting the driving data are based on various technologies, such as Global Navigation Satellite Systems (GNSS), Global Positioning Services (GPS), In-vehicle sensors data extracted from Controller Area Network (CAN) bus, vision systems, and recently utilizing the smartphone's sensors \cite{Seth2020}. We decided to build our IoT-enabled lightweight DBD system to analyze in-vehicle sensor data extracted from the CAN bus in order to compare our data model accuracy on OBD-II compliant and vendor-specific sensor data. OBD-II compliant are used for diagnostics, and addressed using parameter identifiers (PIDs) listed in the SAE J1979 standard. The OBD-II compliant PIDs are similar in all cars produced after 2001 \cite{Lattanzi2021}. The PIDs are used to extract the sensor data by diagnostic scan tools \cite{8239058}. On the other hand, some of the sensors are vendor-specific, and their protocol details are not publicly available. It is a costly task to extract the data for these sensors \cite{Uvarov2021}, which may vary on different car models. In this study, we consider the OBD-II compliant and vendor-specific sensors as public and non-public sensor data, respectively.  

\cite{Lattanzi2021} used SVM and ANN algorithms to classify the drivers’ driving behavior using in-vehicle sensor data. The authors discovered that the accuracy of the proposed algorithms decreased while using only the public sensor data. However, the combination of both public and non-public signals increases the accuracy. \cite{Uvarov2021} also evaluated the performance of different algorithms such as Random Forest, Gradient Boosting, SVM, Decision Tree, and KNN on both public and non-public sensor data. The results show that using only public data decreases the accuracy of proposed classification models. Although Using both types of data obtain good accuracy, it limits the practical use of the existing methods in many applications \cite{Lattanzi2021, Uvarov2021}. For this reason, we made the decision to build our DBD system based on graph convolutional neural networks and long short-term memory networks (GConvLSTM) to increase the accuracy of public signals proposed in \cite{Lattanzi2021}. Additionally, in order to compare the effectiveness of both strategies, we train and assess our model for the combination of public and non-public subsets of sensor data described in the same study. The main contributions of this study are:
\begin{itemize}
\item{We propose a reliable IoT-enabled DBD system using GConvLSTM to provide a robust solution that can differentiate between safe and unsafe driving behavior based on the in-vehicle sensors data extracted from the CAN bus through the OBD-II connector.}
\item{We propose to enhance the accuracy of unsafe driving detection presented in \cite{Lattanzi2021} for public signals.}
\item{We propose to deploy our DBD system at the edge to process and analyze the driving behaviors locally.}
\item{We propose to develop a dashboard that enables the driver to monitor the prediction results, in-vehicle sensor data, and daily reports of driving conditions. Additionally, the dashboard alerts drivers of unsafe driving conditions through voice notification.}
\end{itemize}

\section{Background}
\label{sec:background}
\subsection{Controller Area Network and On Board Diagnostics}
Nowadays, cars are equipped with various electronic control units (ECUs). These ECUs are microcomputers connected to multiple sensors around the car to monitor, control, and optimize the car systems. The ECUs communicate through the CAN, a revolutionary standard communication bus introduced by ROBERT BOSCH GmbH in the early 1980s \cite{Bosch1991}. When any ECU broadcasts data packets over the CAN bus, other ECUs can check the broadcasted packets and decide whether to ignore or receive them. Data extraction directly from the CAN bus for diagnostics and reporting is difficult because all the sensitive data transferred through the CAN bus and interaction with them could be risky and very complicated.  For this reason, higher abstraction level protocols like OBD (On Board Diagnostics) were introduced to the market \cite{birnbaum2001getting} to serve as a self-diagnostic system in vehicles. The OBD works on top of the CAN standard, and the generated OBD messages are encapsulated in CAN messages. In the automotive field, OBD is the main contributor in carrying information and logs exchanged among the ECUs. The OBD-II is the second generation of the OBD protocol, which introduced the universal OBD-II connector for self-diagnostics, reporting and analysis. The OBD scanner is connected to the OBD-II connector to interact with ECUs through reading and writing OBD messages on the CAN bus. Although any car with an OBD-II connector provides access to the data packets generated by ECUs, external diagnostic tools cannot monitor all of them. The users only have access to a subset of monitoring signals such as car speed, engine load, fuel, air pressure, engine revolution speed, etc. The signals from OBD-II are identified by PIDs (parameter identifiers) described in the SAE J1979 standard. The PIDs listed in SAE J1979 are public data and are common in most cars. Some car manufacturers specify additional PIDs to their cars to provide access to more sensors such as steering angle, brake pressure, wheel speeds, etc. Although these signals provide more information about the car's performance, they may vary for different cars, and their protocol details are not publicly available. Hence, using non-public data limits the practical use of the DBD systems. In this study, we aim to increase the accuracy of signals with public PIDs using graph neural networks (GNN), which can make our DBD system more reliable and efficient than previously proposed systems. In addition, we also train and evaluate the proposed GNN model using both public and non-public signals to compare their impact on classification accuracy.
\subsection{Deep Learning and Edge Computing}
Delivering the data generated from IoT devices to the cloud servers for processing is a common practice in the IoT space. However, any delay in exchanging data between the cloud servers and IoT devices can produce unacceptable results in latency-sensitive IoT applications such as autonomous driving, smart healthcare systems, robotics, connected vehicles, etc. The bandwidth cost, latency, and privacy are the challenges that make the cloud computing paradigm inefficient in processing and analyzing the data involved in IoT applications \cite{7331236, 7744809}. Edge computing has been introduced to tackle these challenges efficiently \cite{Vailshery2021}. Offloading computational tasks from cloud data centers to the network edge reduces latency, network traffic, and computational cost in IoT projects. 
Deep learning, on the other hand, is crucial in IoT applications. It outperformed conventional machine learning algorithms in mining raw data generated by devices deployed in complex and noisy IoT environments \cite{7900337}. Therefore, several studies proposed approaches to implement deep learning methods at the edge. \cite{Yao_Tang_Wei_Zheng_Li_2019} deployed a novel spatial-temporal dynamic network (STDN) for traffic forecasting at the edge server using CNN and Long Short-Term Memory (LSTM) networks. \cite{Yuce2017} implemented a neural network-based model at the edge server to analyze the energy consumption of a building, and \cite{Liuarticle} deployed a deep learning-based food recognition system at the network edge to improve the response time. An edge computing strategy can considerably boost the performance of IoT devices, according to prior studies. In order to increase project efficiency, we chose to implement the proposed GNN-based DBD system at the network edge.
\section{Related Work}
\label{sec:related work}
Several approaches have been proposed to differentiate between safe and unsafe driving behaviors to improve road safety. This section highlights the methods used for the driving behavior detection task. \cite{Lattanzi2021} proposed an objective methodology to distinguish between safe and unsafe driving behaviors using the SVM and ANN algorithms based on the data extracted from the CAN bus. The SVM and a feedforward neural network have been trained and evaluated over a publicly available dataset containing more than 26 hours of total driving for ten different drivers. Experimental results show an average accuracy of 90\% for the proposed model in identifying unsafe driving conditions. \cite{Monselise2022} proposed an approach to detect the distinct patterns of aggressive driving based on vehicle sensor data and metadata of each trip. Using the KNN-based model resulted in discovering 4 different aggressive patterns inside the dataset, such as stop-and-go driving, abruptly changing lanes, driving too fast for road conditions, and not properly maintaining lane (or zigzagging) with an average accuracy of 81\%. \cite{Xu2022} utilized HMM and attention-based LSTM networks to predict aggressive driving behavior based on multivariate-temporal feature data such as driver characteristics, environment, and in-vehicle sensor data. Using the proposed model resulted in an average accuracy of 80\%. \cite{9090320} built a deep learning model through stacked denoising sparse autoencoders (SdsAEs) for abnormal driving detection based on the normalized driving behavior data. The proposed SdsAEs model obtained a good performance for detecting abnormal deriving with an accuracy of 98.33\%. \cite{9427166} utilized CNN and LSTM networks to build driving behavior detection models and compare them based on the data extracted from actual two vehicles (V1 and V2). The driving behaviors have been divided into three rankings (A, B, C) for this study, where A, B, and C indicate merit, pass and fail, respectively. The achieved prediction percentages of CNN and LSTM for all driving behavior rankings are higher than 80\%.

Some other studies analyzed the driver's driving behavior for driver identification based on in-vehicle sensor data. \cite{DiGiacomo2021} implemented a study to distinguish an impostor from a car owner through five classifiers; J48Consolidated, RandomTree, J48graft, RepTree, and J48 provided by Weka. Based on the results, J48 and J48graft had better performance in terms of Precision and Recall in identifying the correct class (owner or impostor) for feature vectors belonging to each driver extracted from CAN bus data. \cite{Uvarov2021} compared the impact of using the sensors with/without the non-public PIDs on the performance of the driver identification task. The achieved result indicates that the accuracy of models trained with public parameters decreased by nearly 15\%, where the combination of both public and non-public signals increased the accuracy for all algorithms. Furthermore, the random forest obtained higher accuracy than the decision tree, gradient boosting, SVM, and KNN for both public and non-public data.
\cite{Mekki2019} used fully convolutional networks (FCNs) and LSTM for driver profiling and identification to enhance the security in connected cars. The FCN-LSTM gained higher accuracy than other models after testing on different datasets.
Similarly \cite{Zhang2019a} implemented an end-to-end deep learning framework using CNNs and RNNs with additional attention mechanisms. The in-vehicle sensor data is fed into convolutional layers where the outputs are used as input for recurrent layers with an attention mechanism for time series extraction. Finally, the output layer after recurrent layers is employed to determine the possibility of class distribution for DBD. The DeepConvGRU-Attention and DeepConvLSTM-Attention obtained higher accuracy than existing DeepConvGRU and DeepConvLSTM without attention units.

Most of these studies managed to make a DBD system with good detection accuracy while sacrificing the practical use of the proposed models. As mentioned earlier, using the sensors with/without non-public PIDs affect the accuracy and practical usage of the classification models. In addition, previous studies show that deep learning methods offer a lot of promise for implementing a DBD system based on time series data. In this study, we focus to enhance the practical applicability of our deep learning model by increasing the accuracy of sensors with public PIDs. Furthermore, we investigate to what extent the accuracy of our model can be increased by adding non-public data in the process.
\section{Methodology}
\label{sec:methodology}
The purpose of this study is to design and develop an IoT system that can locally detect the unsafe driving behavior based on the CAN bus data using edge computing and the GConvLSTM algorithm. The output of this study is a DBD system that can be deployed at the edge, where it can process and analyze the drivers' behavior locally to warn them of any unsafe driving conditions. Providing a reliable and efficient DBD system using in-vehicle sensors depends on various factors as follows:

\textbf{Significant feature selection:} Each sensor is a physical entity in our IoT-enabled DBD system and has a column in the driving data set. Thus, the columns in the dataset are virtual entities. Columns are called features of a dataset. The values of these features are used in analyzing driver behaviors. Although all features can be used for DBD, selecting the most significant and relevant features improve the accuracy in classification models and reduce the training time \cite{Zebari2020}. We use 3 subsets of features to evaluate our deep learning model. 2 subsets are similar to selected features in \cite{Lattanzi2021}, and the third subset contains 30 features that are positively or negatively correlated.

\textbf{Classification technique:} In order to classify the driving behaviors, we can use different approaches such as rule-based programming, statistical model, machine learning, or deep learning algorithms. Deep learning algorithms achieved better performance in terms of accuracy and adaptation than other approaches \cite{Chen2019}. In this study, We decided to use graph convolutional recurrent neural networks to classify driving behaviors into safe and unsafe classes, because they have recently achieved high quality and accuracy in spatio-temporal predictions.

\textbf{Deployment environment:} DBD system as an end-to-end IoT solution could be deployed using different layers of computing; cloud, fog, or edge. Compared to cloud computing where the computation services and resources are centralized at large data centers, fog, and edge computing are involved in bringing the computation units closer to the data sources. 
The edge computing approach processes the data such as sensor data close to the logical edge of the network. Some studies like \cite{Mekki2019, Ullah2020} deployed the DBD system at the edge layer to improve response time, security, privacy, and save bandwidth. In this work, we also attempt to deploy our DBD system using an edge computing approach.
\subsection{Dataset}
We use the OCSLab driving dataset \cite{Kwak2016} in this study. This dataset contains 94,401 records with 51 features collected from in-vehicle sensors every second. Each record has been labeled with alphabet letters from “A” to “J”. The size of the dataset is 16.7MB in total. They extracted the driving data using the OBD-II connector from a KIA Motors Corporation car in South Korea while the drivers were driving the car. The number of drivers who participated in this experiment is ten. The length of the driving paths is 23 km and includes three types of motorway, city way, and parking space. All drivers completed 2 round trips for accurate and reliable classification.
\subsection{Data Labeling}
\begin{figure}
\centering
\includegraphics[width=8 cm, trim=90 30 90 70 ,clip]{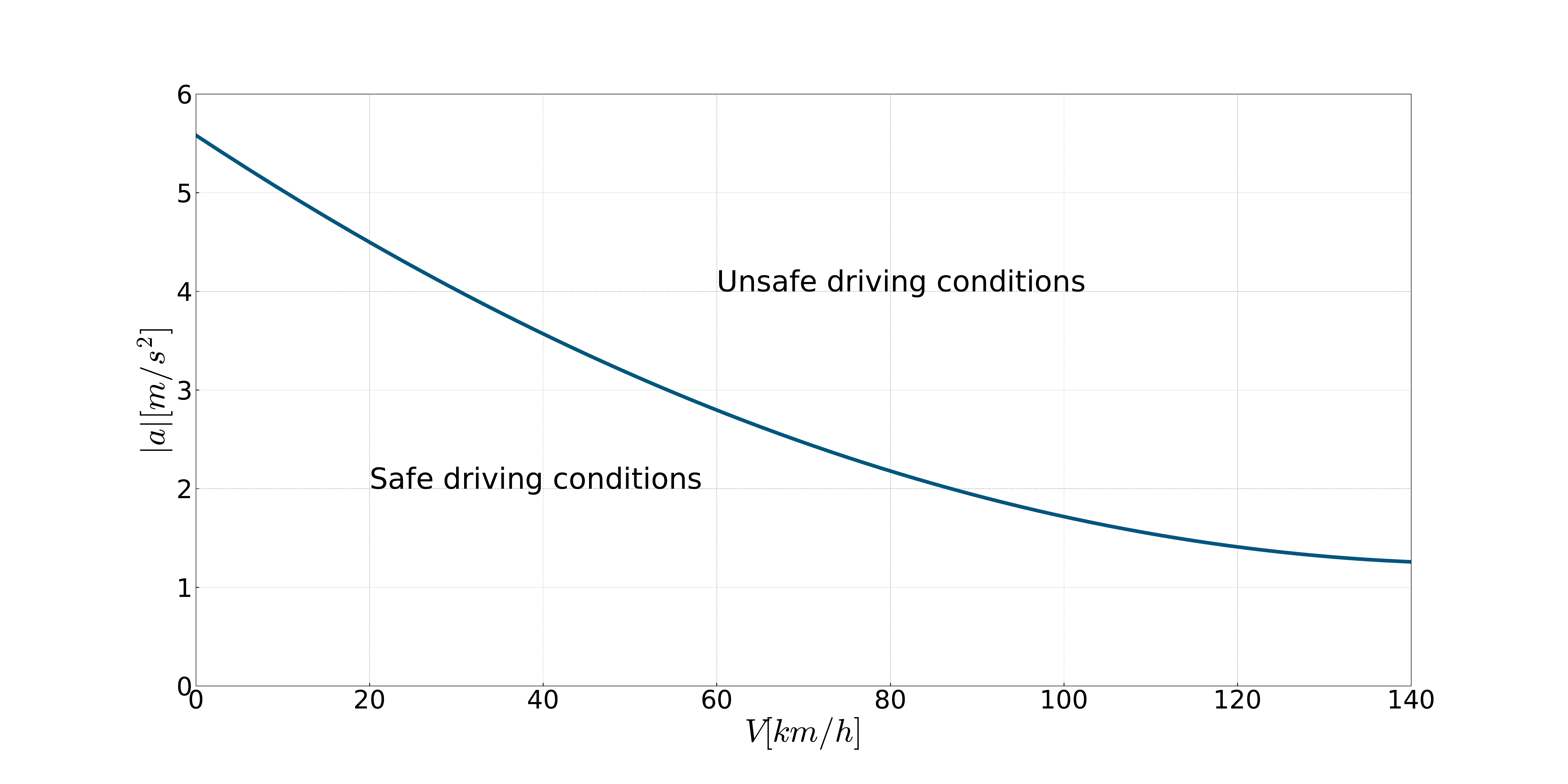}
\caption{The safe and unsafe driving conditions divided using Equation \ref{accel_speed} \cite{Lattanzi2021}.}
\label{fig_1}
\end{figure}

Each supervised classification algorithm requires data associated with proper labels to be used in model training and evaluating tasks. Records in OCSLab driving dataset have been labeled with alphabet letters from “A” to “J”. However, our DBD system aims to differentiate between safe and unsafe driving behavior. For this reason, we need to label the data set again. In order to label the records of the dataset based on our system objectives, we followed the work presented in \cite{Lattanzi2021}. The Equation \ref{accel_speed} has been used in that study to mark each time window with safe and unsafe labels. 
\begin{equation}
\label{accel_speed}
|\bar{a}_{Max}| = g.[0.198 \times (\dfrac{V}{100} )^{2} - 0.592 \times ( \dfrac{V}{100} ) + 0.569],
\end{equation}
which describes a quadratic relationship between acceleration and speed that demonstrates a tolerated acceleration decrease when speed increases \cite{Lattanzi2021}. Figure~\ref{fig_1} shows how this function separates the safe and unsafe areas from each other on the (V, $\vert$ā$\vert$) plane. The points above the curve are considered unsafe driving areas, and below the curve represent the safe driving area. We also use the same approach to label the records in the dataset using Equation \ref{accel_speed}. Then, 69\% of the data have been labeled as “safe” and the rest 31\% as "unsafe".
\subsection{Feature Normalization}
The numerical features of the dataset have different ranges of values, and they need to be normalized. The normalization process changes the value of numerical features to a common scale,  and all the features will be treated equally in the learning algorithm. In addition, it improves the deep learning model performance and stability. The method that we chose is Min-Max scaling which scales the data into a range of [0,1] based on the following equation
\begin{equation}
\label{featnorm}
X_{ i } = \frac{( x_{i} - min(x_{i}) )}{(max(x_{i}) - min(x_{i})  )},
\end{equation}
where min is the minimum value and max is the maximum value of the feature $x_{i}$, respectively.
\subsection{Graph Structured Data}
The OCSLab driving dataset has been constructed from rows and columns. Rows are the observations recorded every second and the columns represent their features. On the other hand, our DBD system is built based on graph neural networks, which only accept graph-structured data. The Graphs are constructed using nodes and edges. We consider the sensors and relationships between them as nodes and edges, respectively. In order to compute the relationships among the in-vehicle sensors, we use the same approach presented in \cite{wangarticle}. Each weighted graph has been represented using a weighted adjacency matrix constructed based on Pearson correlation coefficients (PCCs). They used the PCCs to compute the relationship between the sensors, and it is formulated as
\begin{equation}
\label{pearson}
\rho_{X,Y} = \dfrac{\mathrm{Cov}(X,Y)}{\sigma X \sigma Y},
\end{equation}
Where $\sigma X$ and $\sigma Y$ are standard deviations of $X$ and $Y$, $\mathrm{Cov}$($X$,$Y$) is the covariance of X and Y. The computed $\rho X, Y$ is a value in a range of [-1,1] that represents the relationship between two variables. Then the weighted adjacency matrix is expressed as  
\begin{equation}
\label{adj}
A_{ij} = e^{\rho x_{i}x_{j}},
\end{equation}
\subsection{Proposed Framework}
\subsubsection{Problem Formulation}
\begin{figure}
\centering
\includegraphics[width=8 cm, trim=0 0 0 0, clip]{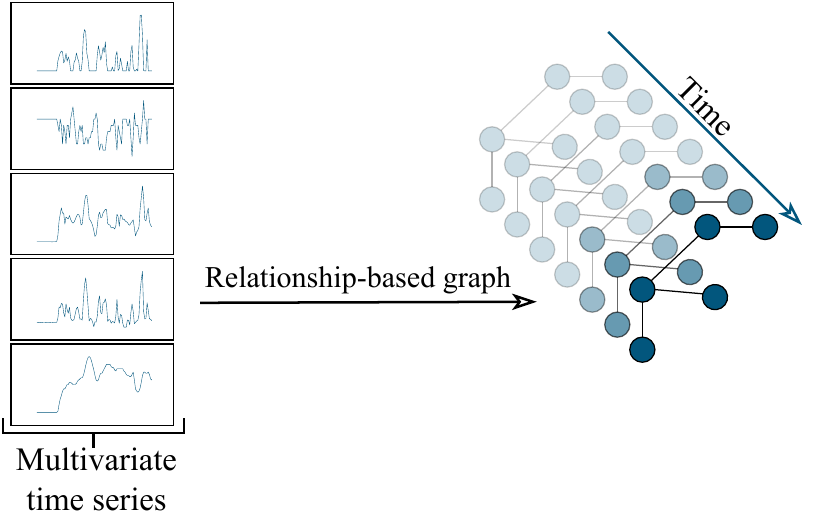}
\caption{Relationship-based graph for observations at time t.}
\label{fig_2}
\end{figure}

This paper focuses on implementing the multivariate time series classification. Given samples of time series observations $S = \{S_{1}, S_{2}, …, S_{N}\}\in\mathbb{R}^{N}$, in which $N$ is the number of time series samples, and corresponding labels $Y = \{y_{1}, y_{2}, …, y_{N}\}\in\mathbb{R}^N$, the model aims to learn a function $f : S \rightarrow Y$ that map $S$ to $Y$ based on the proposed model. Each sample $S_{i}$ contains time series measurements from in-vehicle sensors $x_{t}$ in a time window of size $T$ associated with a label $y \in \{0, 1\}$ – \{0:safe, 1:unsafe\}. Based on the learned function, the model can predict the label $\hat{y}_{i} \in \{0, 1\}$ for $S_{i}$ at certain time steps after $T$. The $x_{t}$ is the data at time step $t$ defined as $x_{t} = (x_{t}^1, x_{t}^2, …, x_{t}^D)^T$, where $D$ shows the dimensionality of $x_{t}$. According to the justification given in the preceding subsection the $x_{t}$ can be expressed as a weighted graph $\mathcal{G} = (V, E, A)$, in which $V$ is a set of finite nodes $|V| = n$, $E$ represents the set of edges and $A \in \mathbb{R}^{n \times n}$ is a weighted adjacency matrix. Consequently, the $S_{i}$ is a sample of multivariate time series where each time step of the observations is a weighted graph associated with the label $y \in \{0, 1\}$. Figure~\ref{fig_2}  shows the relationship-based graph constructed based on observations at time $t$. We considered a time window of 10 seconds with 50\% overlap for this study. 
\subsubsection{System Design and Data Flow}
Our IoT system consists of two main layers; the Data source layer, and the edge layer. The following describes each layer and its functionalities in detail.

\textbf{Data source layer:} This layer has only one component that is responsible for regenerating a subset of the in-vehicle sensors data selected from the dataset. This component has been developed using Python language, and it has two modules. The first module generates sensor values every second. Another is an MQTT client for publishing the generated values to the MQTT broker, where the data analytics engine can consume the sensor's data. In a real-world scenario, the data should be extracted using an OBD-II scanner in real-time. Because of the limitations in implementing this project, we only use this data generation layer to generate values every second to test the performance of our deep learning model, and the monitoring dashboard features locally.

\begin{figure*}
\centering
\includegraphics[width=15 cm, trim= 0 0 0 45, clip]{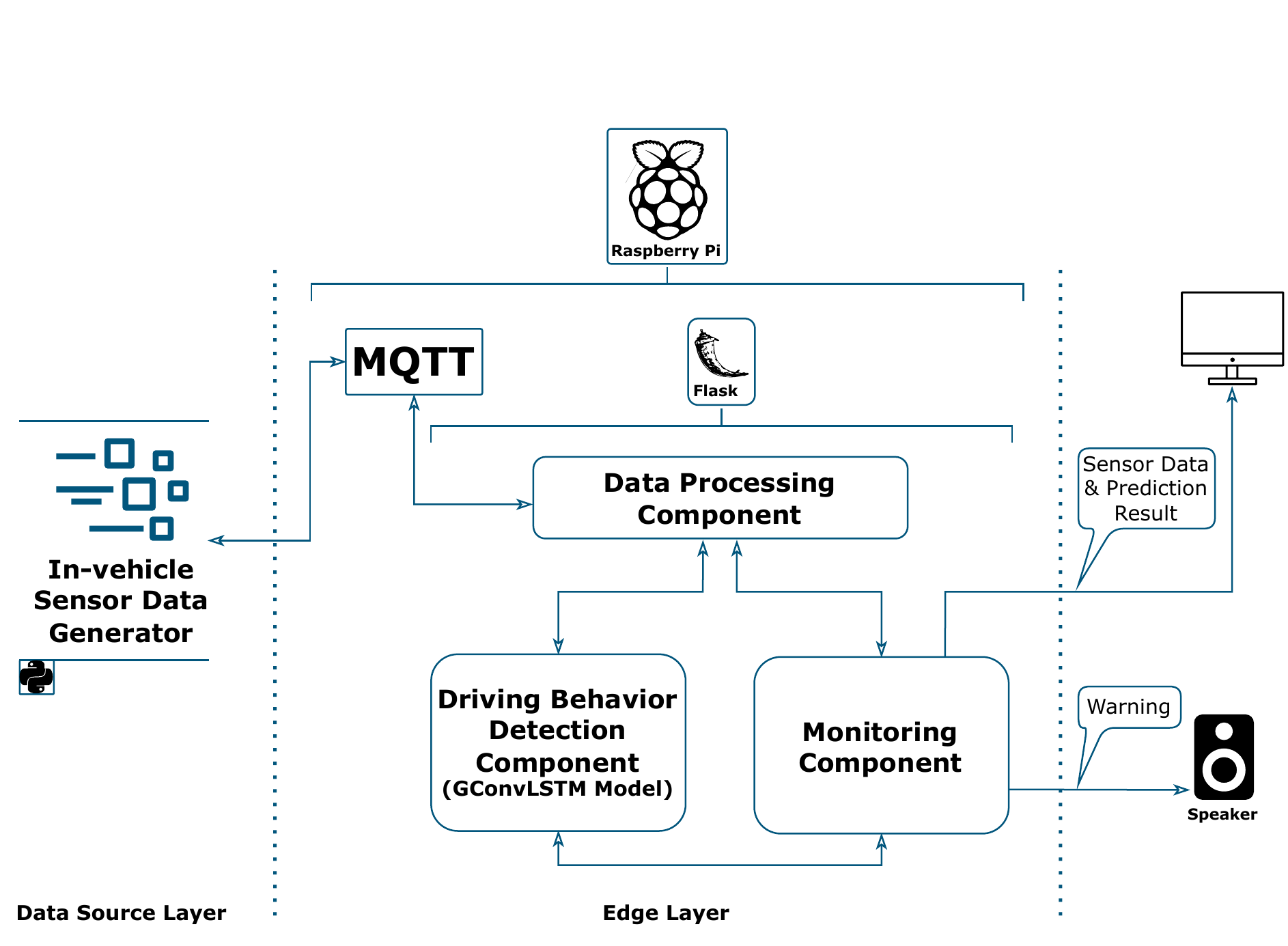}
\caption{Architecture of proposed DBD system.}
\label{fig_3}
\end{figure*}

\textbf{Edge layer:} The edge layer contains three components with specific functionalities. The data processing component provides two services, collecting and preprocessing the sensor data. This component consumes the sensor data from the MQTT broker by subscribing to the predetermined topic. Then, the preprocessing module converts the collected data into graph-structured data. The second component is our DBD model developed based on graph convolutional long short-term memory networks (GConvLSTM) algorithm. The component receives the graph-structured sensor data for the previous ten seconds and classifies the driving behavior for the next ten seconds as “safe” or “unsafe”. The third one is the monitoring component which helps us monitor the in-vehicle sensor data that is updated every second. In addition, the monitoring dashboard alerts the drivers of their unsafe driving behavior through a voice notification. Figure~\ref{fig_3} illustrates the system architecture and data flow for our proposed IoT-enabled DBD system.
\subsubsection{Spatio-temporal Prediction on Graph Data}
In order to implement Spatio-temporal modeling to classify driving conditions, we decided to use GConvLSTM. This is the first study on unsafe driving behavior detection that used the GConvLSTM algorithm. The following sections describe this approach in detail.  

\begin{figure}
\centering
\includegraphics[width=8 cm, clip]{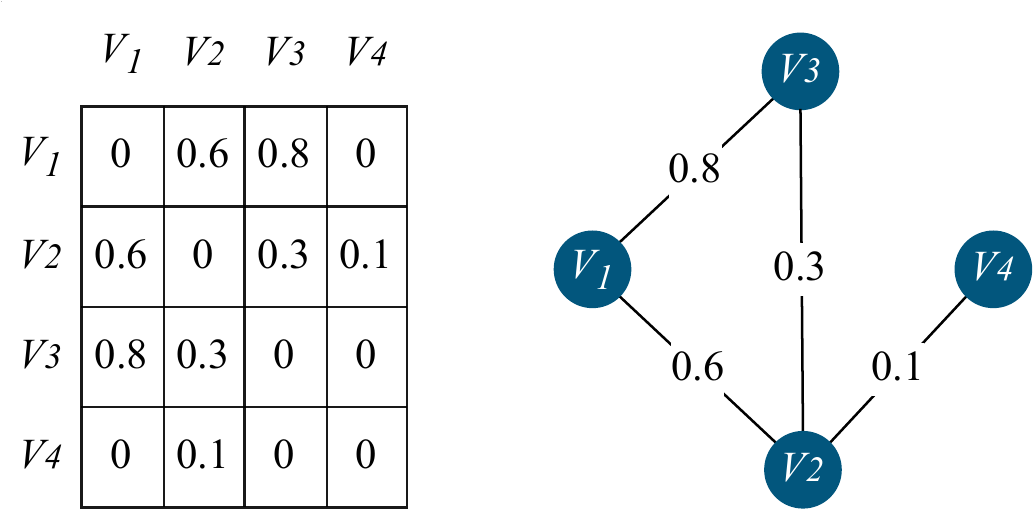}
\caption{Representation of simple weighted undirected graph with its adjacency matrix.}
\label{fig_4}
\end{figure}

\textbf{Graphs:} a graph is a structure that represents a finite set of data objects in which some objects are related to each other. Formally, a graph is expressed as $\mathcal{G} = (V, E)$, where $V$ is a finite set of nodes (objects) and E is a set of edges that shows the relationships between the nodes. Each edge is denoted as $ (u,v)\in E $ which shows the relationship between node $u \in V$ and node $v \in V$. Edge $(u,v) \in E$ can also be written as $e_{uv}$, $uv$ and generally $e$ \cite{9339909}. A convenient and common way to represent a graph is using adjacency matrix $A\in\mathbb{R}^{|V|\times|V|}$. The adjacency matrix is used to order nodes of a graph where each node indicates a particular row and column. If there is an edge between two nodes or a node and itself, we fill the matrix with 1 and 0 otherwise. For instance, the presence of an edge between $u$ and $v$ is expressed as $A[u, v] = 1$ and otherwise $A[u, v] = 0$. The A will be a symmetric matrix if the graph is undirected, this is not required for a directed graph. If the graph contains weighted edges, the entries in the adjacency matrix will be real values rather than {0,1}. Figure~\ref{fig_4} illustrates a simple weighted undirected graph with its adjacency matrix. Graph structured data has rich relation information among objects which can be used to represent complex problems in various areas including social science, natural science, knowledge graphs, and many other domains \cite{ZHOU202057}. The great expressive power of graphs in representing complex problems motivates researchers to apply deep learning methods for graph data.

\textbf{Deep Learning on Graphs:} conventional machine learning approaches require considerable effort and attention to transform the raw data into suitable feature vectors or representations using feature extraction algorithms before applying the learning algorithms \cite{alomarticle}. On the other hand, deep learning methods allow a machine to automatically discover representations from raw data through multiple processing layers that can use non-linear modules \cite{Lecun2015}. The representations are transformed from one layer into another, and the composition of enough transformations can deal with complex data such as time series and graphs \cite{DBLP:journals/corr/abs-2201-00818}. Recently, substantial studies have been implemented to exploit the deep learning methods on ubiquitous graph data which can be categorized as graph convolutional networks, graph recurrent neural networks, graph reinforcement learning, graph encoders, and graph adversarial methods \cite{zhangarticle}. In this study, we focus on graph convolutional and recurrent neural networks to explore multivariate time series classification.

\textbf{Graph Convolutional Networks:} graph convolutional networks (GCNs) can be considered as convolutional neural networks (CNNs) for graph data. Multiple processing layers, local connections, shared weights and pooling layers are key ideas behind the CNNs architecture \cite{Lecun2015}, which allow a machine to extract and compose the spatial features to construct meaningful representations from euclidean data such as images. These operations are also essential in dealing with graph data while inspecting neighboring nodes to discover the representations from one layer to another one. The difference in implementing the CNNs and GCNs is that CNNs are built for Euclidean or regular structured data while the GCNs are designed for non-Euclidean structured data. Hence, the convolutions filters and pooling operations on graphs are not as well-defined on Euclidean data such as images \cite{zhang2article}. The convolutional operations on graph-structured data are categorized as spatial-based GCNs and spectral-based GCNs. We use spectral-based GCNs to implement the proposed driving behavior detection model. 

In spatial-based methods, graph convolutions are defined based on nodes’ spatial relations as analogous to convolutional operations in CNN models on images. The CNN models are designed with some implicit assumptions about the structure of input data. For instance, images are grid-like data where each pixel has a fixed number of neighbors, and the spatial order for scanning the images is naturally determined \cite{zhang2article}. However, in arbitrary graph data, the number of neighbors for each node and the spatial order of the nodes is variable. Some studies have been proposed to perform spatial-based GCNs on graph data \cite{DBLP:journals/corr/NiepertAK16, DBLP:journals/corr/MontiBMRSB16, DBLP:journals/corr/abs-1808-03965}.

The receptive fields used in spatial approach are spatially constructed directly on graphs while the convolution operations in the spectral domain rely on spectral graph theory, where the graph signals need to be transformed into the spectral domain through eigenvectors of graph Laplacian. 
The first framework for spectral convolution on graph data has been implemented through the graph Laplacian matrix proposed by \cite{bruna2013spectral}. The important issue of this approach is eigenvalue decomposition which has $O(n^3)$ computational complexity. \cite{defferrard2016convolutional} facilitated this drawback by offering linear complexity $O(|E|)$ and providing a strictly localized filter. \cite{DBLP:journals/corr/KipfW16} developed a novel model for semi-supervised classification on graphs based on first-order approximation of spectral convolutional filters proposed by Defferrard. In this study, we use a model proposed by \cite{seo2018structured} based on the Defferrard technique, where the spectral formulation of convolution operation is defined in the Fourier domain as 
\begin{equation}
\label{convopt2}
x*\mathcal{G} y = U(( U^{T} x) \odot (U^{T} y)),
\end{equation}
where $\odot$ is an element-wise Hadamard product. It follows that a signal $x$ is filtered by $g_{\theta}$ as 
\begin{equation}
\label{filter}
y = g_{\theta}( L )x = g_{\theta}( U\Lambda U^{T} )x = Ug_{\theta}(\Lambda)U^{T}x,
\end{equation}
where $g_{\theta}(\Lambda) = diag(\theta)$ is a non-parametric kernel and $ \theta \in R^n$ is a vector of Fourier coefficients.

The normalized graph Laplacian is a matrix representation of a graph and it is defined as $ L = I_{n} - D^{-{1}/{2}} AD^{-{1}/{2}} $, in which $ I_{n} $ is the identity matrix and $ D \in \mathbb{R}^{n \times n} $ is the diagonal matrix of node degrees with $ D_{ii} = \sum_{j} (A_{i,j}) $ \cite{seo2018structured}. The normalized graph Laplacian matrix can be diagonalized through orthogonal matrix $U$ such that $L=U \Lambda U^T$, where $U = [u_{0}, u_{1}, …, u_{n-1}] \in \mathbb{R}^{n \times n}$ is a matrix of eigenvectors and $\Lambda \in \mathbb{R}^{n \times n}$ is the diagonal matrix of eigenvalues of the normalized graph Laplacian. The graph signal $x$ is filtered through the element-wise multiplication of $g_{\theta}$ and $U^T x$, which is the graph Fourier transform of $x$. However, the learning complexity of a non-parametric filter is in $ O(n) $, and computing the eigenvalue decomposition of $L$ for large graphs can be expensive. To overcome this issue, Defferrard  parameterized the filter $ g_{\theta} $ by Chebyshev polynomials $T_{k}$ expressed as
\begin{equation}
\label{cheby}
g_{\theta}(\Lambda)= \sum_{K=0}^{K-1} \theta_{k}T_{k} ( \tilde{\Lambda} ) ,
\end{equation}
where the parameter $\theta \in R^K$ is a vector of Chebyshev coefficients and $T_{k}(\tilde{\Lambda}) \in \mathbb{R}^{n \times n}$ is the Chebyshev polynomial of order $k$ evaluated at $\tilde{\Lambda}= 2\Lambda/\lambda_{max}-I_{n}$, the value of $\tilde{\Lambda}$ is in range of [-1, 1]. The graph filtering operation can be defined as
\begin{equation}
\label{filter2}
y = g_{ \theta } * \mathcal{G}  x = g_{ \theta }( L )x= \sum_{k=0}^{K-1} \theta_{ k }T_{ k }(  \tilde{L})x,  
\end{equation}
where $ T_{k}(\tilde{L}) \in \mathbb{R}^{n \times n} $ is the Chebyshev polynomial of order $k$ evaluated at the scaled Laplacian $\tilde{L} = 2L / \lambda_{max} - I_{n}$.

\textbf{Long Short-Term Memory Network (LSTM):} LSTM introduced by \cite{hochreiter1997long} and it is a special type of recurrent neural network (RNN) that is extensively used in time series forecasting. It is difficult for RNNs to carry information to current time steps from earlier ones, which leads to information loss. In order to keep the influence of earlier inputs on the prediction, the LSTM adds a unit to RNN for storing the long-term states, which are managed through the gates. As a result, the LSTM helps in avoiding the vanishing gradient problem. In this study, we follow the LSTM variation described by \cite{DBLP:journals/corr/Graves13} and a LSTM cell is formulated as
\begin{equation}
\label{lstmformula1}
\begin{gathered}
i= \sigma ( W_{xi}X_{t} + W_{hi}h_{t-1} + W_{ci} \odot C_{t-1} + b_{i}),\\
f = \sigma ( W_{xf}X_{t} + W_{hf}h_{t-1} + W_{cf} \odot C_{t-1} + b_{f}),\\
C_{ t } = f_{ t } \odot C_{ t-1 } + i_{ t } \odot tanh( W_{xc}X_{t} + W_{hc}h_{t-1} + b_{c} ),\\
o = \sigma ( W_{xo}X_{t} + W_{ho}h_{t-1} + W_{co} \odot C_{t} + b_{o}),\\
h_{ t } = o \odot tanh( c_{t} ), 
\end{gathered}
\end{equation}
each cell of the LSTM unit is considered a memory unit with a state $C_{t}$ at time $t$. The $f$, $i$ and $o$ indicate the forget gate, input gate and output gate, respectively. Reading or modifying this memory unit is performed through these gates. In a nutshell, they learn and decide which information in a sequence is essential to be kept or discarded. At each time step, the LSTM cell receives inputs from two sources: the current input $x_{t}$ and previous hidden states $h_{t-1}$. The weights are represented by $W_{x} \in \mathbb{R}^{d_{h} \times d_{x}} $,  $W_{h} \in \mathbb{R}^{d_{h} \times d_{h}} $,  $W_{c} \in \mathbb{R}^{d_{h}} $ and biases are shown as $ b_{i}, b_{f}, b_{c}, b_{o} \in \mathbb{R}^{d_{h}}$ . In the formula, $\odot$ is the Hadamard element-wise product and $\sigma (.)$  denotes the sigmoid function. This LSTM model use Peephole connections introduced by \cite{inproceedingsPeephole} which means that each gate has a peephole connection with the cell state.

\textbf{Convolutional LSTM (ConvLSTM):} ConvLSTM is a form of LSTM for spatio-temporal prediction introduced by \cite{shiarticle} which replaces internal matrix multiplications with convolution operations at layer transitions. The following are the key equations of ConvLSTM, where $*$ denotes the convolution operator and $\odot$ the Hadamard product
\begin{equation}
\label{lstmformula2}
\begin{gathered}
i= \sigma ( W_{xi}*X_{t} + W_{hi}*h_{t-1} + W_{ci} \odot C_{t-1} + b_{i}),\\
f = \sigma ( W_{xf}*X_{t} + W_{hf}*h_{t-1} + W_{cf} \odot C_{t-1} + b_{f}),\\
C_{ t } = f_{ t } \odot C_{ t-1 } + i_{ t } \odot tanh( W_{xc}*X_{t} + W_{hc}*h_{t-1} + b_{c} ),\\
o = \sigma ( W_{xo}*X_{t} + W_{ho}*h_{t-1} + W_{co} \odot C_{t} + b_{o}),\\
h_{ t } = o \odot tanh( c_{t} ),
\end{gathered}
\end{equation}

\textbf{Graph Convolutional LSTM (GConvLSTM)}: As previously stated, traditional convolutional operations are unsuitable for non-Euclidean structured data. We use the graph convolutional model proposed by \cite{seo2018structured} in order to implement multivariate time series classification on graph data. They generalized the ConvLSTM model presented by \cite{shiarticle} for graphs data through replacing the 2D convolution operation $*$ by the graph convolution operation $*\mathcal{G}$ 
\begin{equation}
\label{GConvlstmformula}
\begin{gathered}
i= \sigma ( W_{xi} \, {*\mathcal{G}} \, X_{t} + W_{hi} \, {*\mathcal{G}} \, h_{t-1} + W_{ci} \odot C_{t-1} + b_{i}),\\
f = \sigma ( W_{xf} \, {*\mathcal{G}} \, X_{t} + W_{hf} \, {*\mathcal{G}} \, h_{t-1} + W_{cf} \odot C_{t-1} + b_{f}),\\
C_{ t } = f_{ t } \odot C_{ t-1 } + i_{ t } \odot tanh( W_{xc} \, {*\mathcal{G}} \, X_{t} + W_{hc} \, {*\mathcal{G}} \, h_{t-1} + b_{c} ),\\
o = \sigma ( W_{xo} \, {*\mathcal{G}} \, X_{t} + W_{ho} \, {*\mathcal{G}} \, h_{t-1} + W_{co} \odot C_{t} + b_{o}),\\
h_{ t } = o \odot tanh( c_{t} ), 
\end{gathered}
\end{equation}
This model is an implementation of GConvLSTM, where the graph convolutional kernels are defined using Chebyshev coefficients as $ W_{x} \in \mathbb{R}^{K \times d_{h} \times d_{x}} $, $ W_{h} \in \mathbb{R}^{K \times d_{h} \times d_{h}}$, the $K$ indicates the number of parameters which is independent from number of nodes $n$. The $ W_{xi} \, {*\mathcal{G}} \, x_{t}$ performs the graph convolution operation on $x_{t}$ using $d_{x}d_{h}$ filters which are the functions based on graph Laplacian $L$ parameterized by $K$ Chebyshev coefficients as mentioned in Equations \ref{cheby} and \ref{filter2}.  
\section{Experiments and Results}
\label{sec:experiments and results}
\subsection{GConvLSTM}
\begin{figure*}[!t]
\centering
\includegraphics[width=15 cm, trim=0 0 0 0, clip]{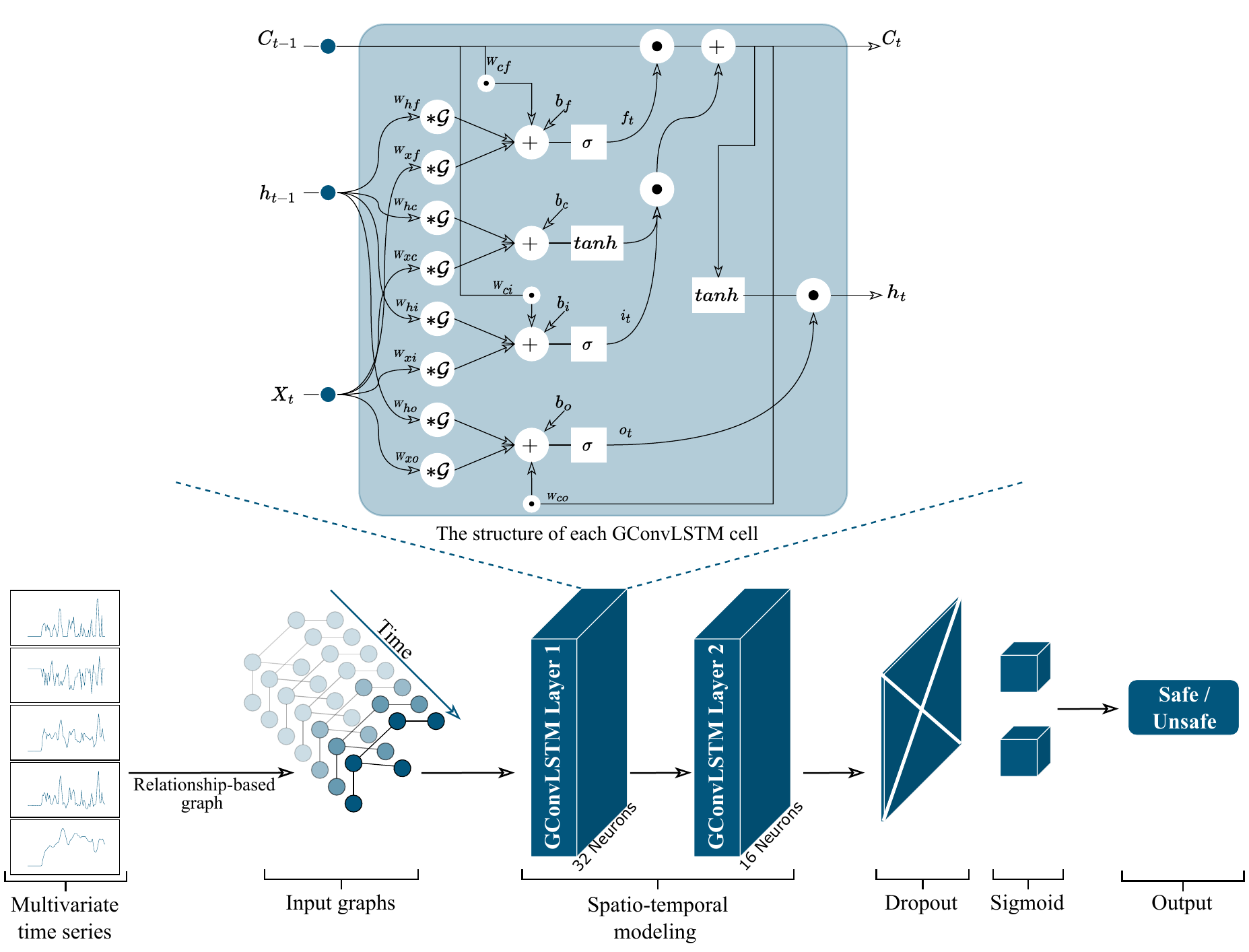}
\caption{The overview of implemented GConvLSTM model in proposed DBD system.}
\label{fig_6}
\end{figure*}

We used GConvLSTM to implement unsafe driving behavior detection. The input of the GConvLSTM model is 10 time steps (10 seconds) of in-vehicle sensors data, and the output of the model is a label/class predicted for the next 10 time steps. The predicted label/class can be \{0:”safe driving condition” \} or \{1:”unsafe driving condition”\}. We considered two hidden layers for our GConvLSTM model, where the first hidden layer has 32 neurons and the second hidden layer contains 16 neurons. The dropout technique is implemented before the output layer to prevent the model from overfitting. The probability value for dropout is 0.5. Dropout randomly sets the elements on the input layer to zero based on the specified probability only during the training stage to decrease overfitting. The output layer labels the representations produced by the previous layer through the sigmoid activation function. The model is trained by minimizing the binary cross-entropy loss (BCELoss) and optimizing through the Adam optimizer. The loss function helps to evaluate the candidate set of weights which leads the model to learn to decrease the error in prediction. The BCELoss is expressed as
\begin{equation}
\label{BCELoss}
BCELoss = \dfrac{1}{N} \sum_{i=1}^{N}-( y_{i} \times log( p_{i}) + (1-y_{i}) \times log(1-p_{i})), 
\end{equation}
where, $y$ is the label (0 or 1), $p_{i}$ is the probability of class 1, and $1-p_{i}$ is the calculated probability of class 0. The Adam optimization method is considered an extension for stochastic gradient descent to update the weights of the network iteratively in training time. Adam's method is effective and efficient because it can achieve good results and needs less memory for the problem with a huge amount of data. In this study, the learning rate for the Adam optimizer is 0.001 and the number of epochs to train the model is 20. Figure~\ref{fig_6} shows the structure of the GConvLSTM model.
\subsection{Evaluation}
The training dataset contains 80\% of the information, and the rest 20\% of the driving dataset is assigned as the test dataset. We used the training dataset to train the model and the test dataset to validate the model. The performance of the developed GConvLSTM model has been evaluated through the accuracy, precision, recall, and F1-score metrics expressed as 
\begin{equation}
\label{accuracy1}
Accuracy =  \dfrac{(TP+TN)}{(TP+TN+FP+FN)},
\end{equation}
\begin{equation}
\label{accuracy2}
Precision = \dfrac{TP}{(TP+FP)}, 
\end{equation}
\begin{equation}
\label{accuracy3}
Recall = \dfrac{TP}{(TP+FN)}, 
\end{equation}
\begin{equation}
\label{accuracy4}
F1-score = \dfrac{2 \times ( Precision \times Recall )}{(Precision + Recall)}, 
\end{equation}
where $TP$ (true positive) shows the number of positive samples classified accurately, $TN$ (True Negative) indicates the number of negative samples that are classified accurately, $FP$ (False Positive) indicates the number of negative samples that are classified as positive and $FN$ (False Negative) shows the number of positive samples that classified as negative. 
According to previous studies, using only sensors non-public PIDs decreases the accuracy of the classification algorithms. However, the combination of both public and non-public signals increases the accuracy \cite{Lattanzi2021, Uvarov2021}. Although utilizing the signals without public PIDs enhances the performance of classification models, it limits their practical usage. In this study, we focus on improving the practical applicability of our deep learning model by increasing the accuracy of sensors with public PIDs. Furthermore, we investigate the impact of the non-public data on our proposed model. The driving dataset used for this study contains 51 signals. Some signals are not compliant with the OBD-II standard and have non-public PIDs. The public and non-public subsets used for training and evaluating the model are defined in Table~\ref{tab:table1}. Subset A contains public signals, and subset B has both public and non-public signals. Both subsets A and B are same as the selected features used in \cite{Lattanzi2021}. In addition, we used subset C to check to what extent the accuracy of the GConvLSTM model is influenced by non-public signals. Subset C contains all public and non-public features that are positively and negatively correlated.
{\renewcommand{\arraystretch}{1.2}
\begin{table*}[!t]
\caption{Selected features for each subset \label{tab:table1}}
\setlength{\tabcolsep}{0.1cm}
\centering
\begin{tabular}{|p{4cm}|p{13cm}|}
\hline
Subset A (Public) & Vehicle speed, Engine speed, Engine load, Throttle position.\\
\hline
Subset B (Public+non-Public) & Vehicle speed, Engine speed, Engine load, Throttle position, Steering wheel angle, Brake pedal pressure\\
\hline
Subset C (Public+non-Public) & Fuel consumption, Accelerator pedal value, Throttle-position-signal, Short term fuel trim bank1, Intake-air-pressure, Absolute throttle position, Engine speed, Engine torque after correction, Torque of friction, Flywheel torque (after torque interventions), Current spark timing, Engine coolant temperature, Engine idle target speed', Engine torque, Calculated load value, Flywheel torque, Torque converter speed, Engine coolant temperature.1, Wheel velocity front left-hand, Wheel velocity rear right-hand, Wheel velocity front right-hand, Wheel velocity rear left-hand, Torque converter turbine speed -Unfiltered, Vehicle-speed, Acceleration speed-longitudinal, Master cylinder pressure, Calculated road gradient, Acceleration speed-Lateral, Steering wheel speed, Steering wheel angle\\
\hline
\end{tabular}
\end{table*}
}
\subsection{Edge Server}
We developed our edge server using the Python Flask framework on Raspberry Pi (RPI). The RPI used for this study is an RPI 4 model B with a 64-bit quad-core Cortex-A72 processor and 8GB LPDDR4 RAM. 
\subsection{Results}
\begin{figure}
\centering
\includegraphics[width=8 cm, trim=20 20 20 30, clip]{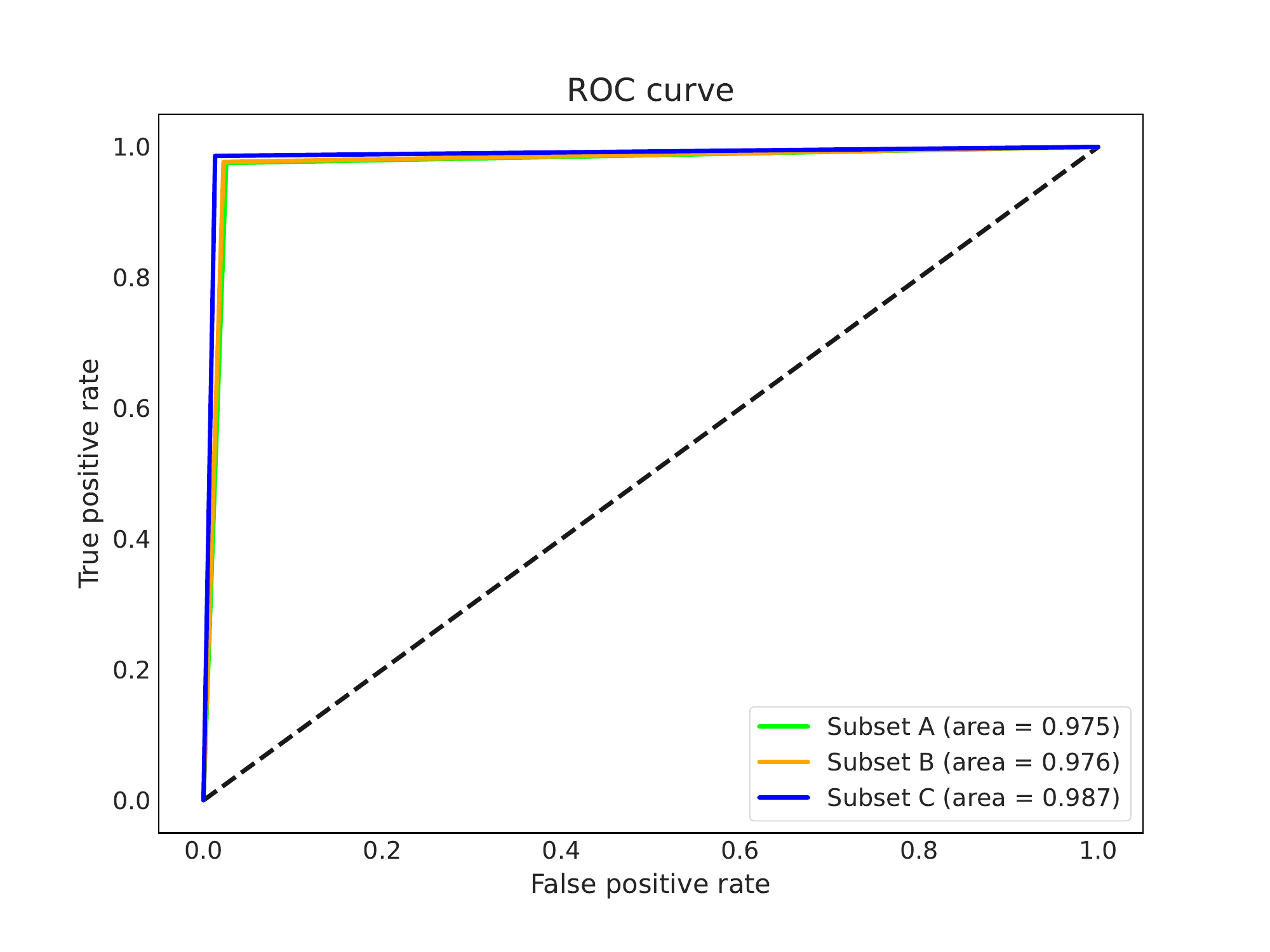}
\caption{ROC curves and AUC scores of GConvLSTM for all subsets.}
\label{fig_7}
\end{figure}

The results show that the GConvLSTM model has a better performance compared to the previous study \cite{Lattanzi2021} in detecting unsafe driving behavior. We only compare our results to this study because we used the same data labeling approach and dataset as they did. Table~\ref{tab:table2} compares the performance of our model with SVM and NN proposed in \cite{Lattanzi2021} on subset A which contains public signals. Our methods developed based on GCovnLSTM achieved the best results for all evaluation metrics. 
{\renewcommand{\arraystretch}{1.2}
\begin{table}[!t]
\caption{Performance comparison of methods on subset A \label{tab:table2}}
\setlength{\tabcolsep}{0.2cm}
\centering
\begin{tabular}{p{1.4cm} p{1.1cm} p{1.1cm} p{1.1cm} p{1.2cm}}
 \hline
 Method & Accuracy & Precision & Recall & F1-Score \\ [0.5ex] 
 \hline
 SVM  & 82.5\% & 82.1\% & 76.3\% & 79.1\% \\ 
 NN  & 84.1\% & 82.9\% & 78.2\% & 80.5\%\\ 
 \textbf{GCovnLSTM}  & \textbf{97.5\%} & \textbf{97.6\%} & \textbf{97.5\%} & \textbf{97.5\%}\\
 \hline
\end{tabular}
\end{table}
}
Table~\ref{tab:table3} shows the performance comparison of the models on subset B, which contains both public and non-public signals. The results show the impact of the non-public signals on classification accuracy for all models. The performance for SVM and NN increased significantly, while there is a slight improvement in our method. Therefore, GConvLSTM is more stable and accurate in driving behavior detection for both public and non-public sensor data.
{\renewcommand{\arraystretch}{1.2}
\begin{table}[!t]
\caption{Performance comparison of methods on subset B \label{tab:table3}}
\setlength{\tabcolsep}{0.2cm}
\centering
\begin{tabular}{p{1.4cm} p{1.1cm} p{1.1cm} p{1.1cm} p{1.2cm}}
 \hline
 Method & Accuracy & Precision & Recall & F1-Score \\ [0.5ex] 
 \hline
 SVM  & 90.2\% & 89.4\% & 87.7\% & 88.5\% \\ 
 NN  & 91.7\% & 90.8\% & 89.1\% & 89.9\%\\ 
 \textbf{GCovnLSTM}  & \textbf{97.5\%} & \textbf{99.6\%} & \textbf{95.8\%} & \textbf{97.6\%}\\
 \hline
\end{tabular}
\end{table}
}
Moreover, we also evaluated our model performance with subset C, which has 30 features. Subset C contains all positively and negatively correlated signals from the driving dataset. Table~\ref{tab:table4} shows the performance results on subset C. Compared to subset A which contains public signals, the accuracy improved by 1.2\%. It is a good achievement. It implies that our model is more stable and reliable on both public and non-public signals. Therefore, our model is more practical than previous models for signals with only public PIDs.
{\renewcommand{\arraystretch}{1.2}
\begin{table}[!t]
\caption{Performance evaluation of GConvLSTM on subset C \label{tab:table4}}
\setlength{\tabcolsep}{0.2cm}
\centering
\begin{tabular}{p{1.4cm} p{1.1cm} p{1.1cm} p{1.1cm} p{1.2cm}}
 \hline
 Method & Accuracy & Precision & Recall & F1-Score \\ [0.5ex] 
 \hline
 \textbf{GCovnLSTM}  & \textbf{98.7\%} & \textbf{98.6\%} & \textbf{98.6\%} & \textbf{98.6\%}\\
 \hline
\end{tabular}
\end{table}
}
Furthermore, we compared our model performance on subsets A, B, and C using ROC (Receiver Operating Characteristic) curve and AUC (Area Under the Curve). Figure~\ref{fig_7} shows the ROC curve for the “unsafe” class for subsets A, B, and C. The AUC scores are very close, and the strong overlap of ROC curves implies that our model can produce stable results for both public and non-public in-vehicle sensor data. 
\subsection{Monitoring Dashboard}
\begin{figure*}
\centering
\includegraphics[width=15 cm]{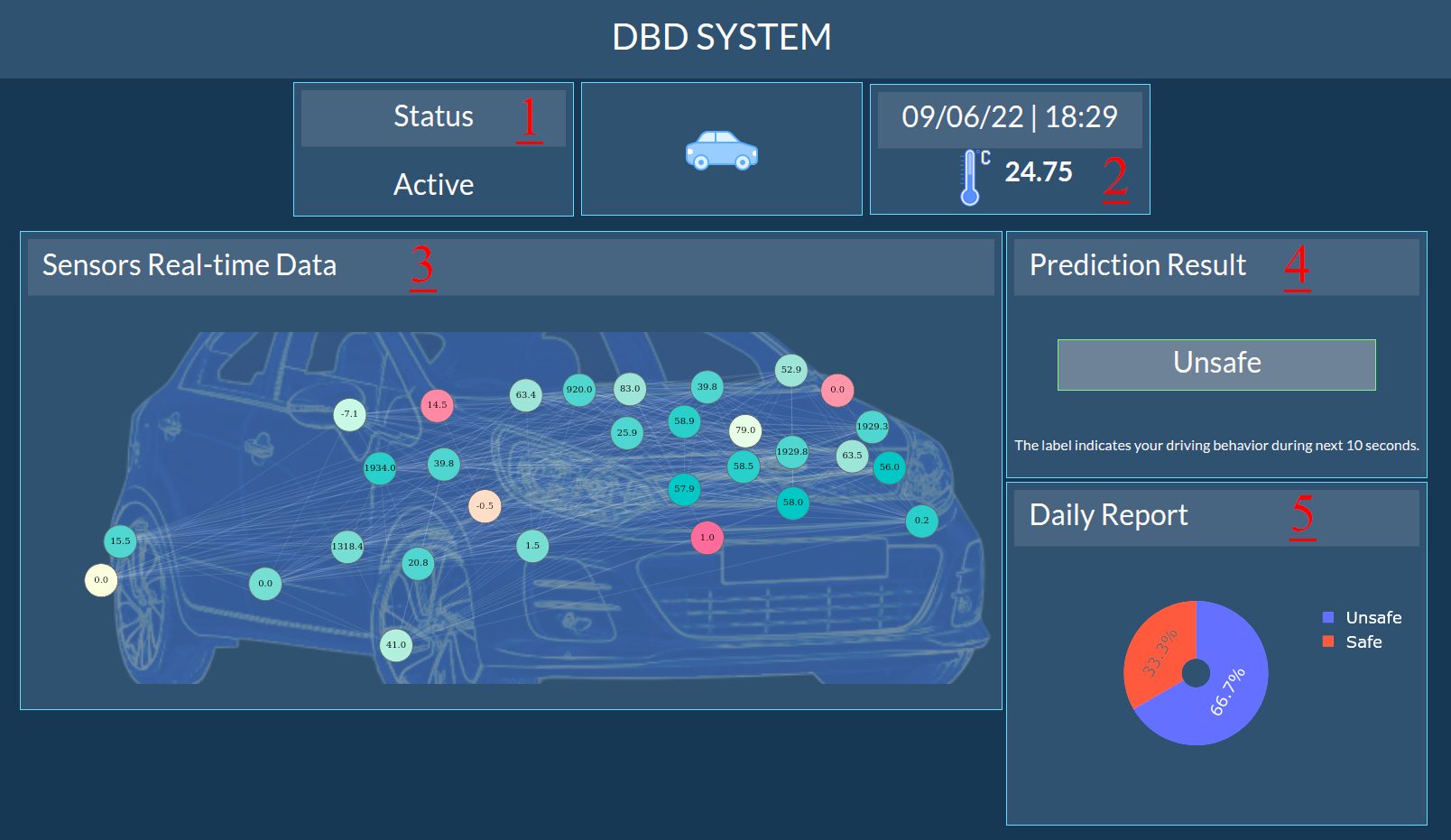}
\caption{Monitoring dashboard user interface.}
\label{fig_8}
\end{figure*}

Figure~\ref{fig_8} shows the monitoring dashboard developed to track the performance of our DBD system. The dashboard has five components built using the Python Dash framework. The first component shows the status of the DBD system. If every part of the system works accurately, the state indicates active. Otherwise, it displays inactive. The second component shows the current date, time, and outside temperature. The third one is the graph representation of the sensor data updated every second. The user can touch or hover each node to get more information about that sensor. The position of the sensors is not accurate, and they are selected randomly. The edges are created based on the calculated correlation of the sensors. The fourth represents the predicted label for the next 10 seconds. The Fifth component is a live pie chart that shows the daily report of predicted labels. The dashboard also alerts the driver using voice notification for any detected unsafe driving behavior.
\section{Conclusion and Future Works}
\label{sec: conclusion and future works}
In this study, we proposed an IoT system capable of detecting unsafe driving behavior using in-vehicle sensor data. We utilized the OCSLab driving dataset, which contains sensor data extracted from the CAN bus via an OBD-II connector, to evaluate our approach. Our proposed model, based on the GConvLSTM algorithm, improved the accuracy of unsafe driving behavior detection using publicly available sensor data, making our approach more practical than existing methods while achieving an accuracy of 97.5\% for public sensors. Furthermore, we investigated the impact of non-public data on the GConvLSTM model, which resulted in a slight improvement. The average accuracy for the combination of public and non-public sensors was 98.1\%, highlighting the reliability and efficiency of our proposed approach for detecting unsafe driving behavior in both scenarios.

We deployed our proposed DBD system on an RPI 4 Model B to enable the local detection of unsafe driving behavior. We also developed a monitoring dashboard that displays sensor data, prediction results, and daily reports on driving behavior conditions. The system alerts drivers via voice notifications of any detected unsafe driving behavior. As a lightweight DBD system, the developed approach can enhance road safety when implemented in vehicles. 

For future research, we plan to create a graph dataset based on the dynamic nature of in-vehicle sensor networks and investigate how this can improve the accuracy and robustness of GConvLSTM in real-world driving behavior detection tasks. We also intend to incorporate additional features into the monitoring dashboard, such as personalized driving behavior recommendations based on individual driving performance, to assist drivers in improving their driving skills.

In conclusion, our proposed IoT system using the GConvLSTM algorithm can effectively detect unsafe driving behavior with high accuracy using publicly available sensor data. The system is lightweight, practical, and can be deployed locally in vehicles to improve road safety. Our research findings can contribute to the development of more reliable and efficient driving behavior detection systems, potentially reducing the incidence of accidents caused by driver errors.


\bibliographystyle{IEEEtran}

\bibliography{./manuscript}  

\end{document}